\documentclass{article}

\usepackage{amsmath}
\usepackage[dvips]{graphicx}
\usepackage{dsfont}
\usepackage{amssymb}
\usepackage{amstext}
\DeclareGraphicsExtensions{.eps}
\usepackage{pdfsync}
\usepackage{subfigure}

\input xy
\xyoption{all}

\usepackage{algorithm}
\usepackage{algorithmic}

\usepackage{proof}

 \newtheorem{prop}{Proposition}

\begin{document}

\title{Reverse Engineering and Symbolic Knowledge Extraction on {\L}ukasiewicz Fuzzy Logics using Linear Neural Networks}

\author{ Carlos Leandro \\ \texttt{[miguel.melro.leandro@gmail.com]} 
\\ \texttt{ Departamento de Matem\'{a}tica,}
\\ \texttt{Instituto Superior de Engenharia de Lisboa, Portugal.}
}

\maketitle

\begin{abstract}
This work describes a methodology to combine logic-based systems and connectionist systems. Our approach uses finite truth valued {\L}ukasiewicz logic, where we take advantage of fact, presented by Castro in \cite{Castro98}, what in this type of logics every connective can be define by a neuron in an artificial network having by activation function the identity truncated to zero and one. This allowed the injection of first-order formulas in a network architecture, and also simplified symbolic rule extraction.

Our method trains a neural network using Levenderg-Marquardt algorithm, where we restrict the knowledge dissemination in the network structure. We show how this reduces neural networks plasticity without damage drastically the learning performance. Making the  descriptive power of produced neural networks similar to the descriptive power of {\L}ukasiewicz logic language, simplifying the translation between symbolic and connectionist structures.

This method is used  in the reverse engineering problem of finding the formula used on generation of a truth table for a multi-valued {\L}ukasiewicz logic. For real data sets the  method is particulary useful for attribute selection, on binary classification problems defined using nominal attribute. After attribute selection and possible data set completion in the resulting connectionist model: neurons are directly representable using a disjunctive or conjunctive formulas, in the {\L}ukasiewicz logic, or neurons are interpretations which can be approximated by symbolic rules. This fact is exemplified, extracting symbolic knowledge from connectionist models generated for the data set \emph{Mushroom} from \emph{UCI Machine Learning Repository}.
\end{abstract}

\section*{Introduction}

There are essentially two representation paradigms, namely, connectionist representations and symbolic-based representations, usually taken as very different. On one hand, symbolic based descriptions is specified through a grammar having a fairly clear semantics, can codify structured objects, in some cases support various forms of automated reasoning and can be
transparent to users. On the other hand the usual way to see information presented using connectionist description, is its codification on a neural network. Artificial neural networks in principle combine, among other things, the ability to learn (and be trained) with massive parallelism and robustness or insensitivity to perturbations of input data.
But neural networks are usually taken as black boxes providing little insight into how the information is codified. They have no explicit, declarative knowledge structure that  allows the representation and generation of explanation structures. Thus, knowledge captured by neural networks is not transparente to users and cannot be verified by domain experts. To solve this problem, researchers have been interested in developing a humanly understandable  representation for neural networks.

It is natural to seek a synergy integrating the \emph{white-box} character of symbolic base representation and the learning power of artificial neuro networks. Such neuro-symbolic model are currently a very active area of research. One particular aspect of this problem which been considered in a number of papers, see \cite{Bornscheuer98} \cite{Hitzler04} \cite{Holldobler00} \cite{Holldobler94} \cite{Holldobler99}, is the extraction of logic programs from trained networks.

Our approach to neuro-symbolic models and knowledge extraction is based on trying to find a comprehensive language for humans representable directly in a neural network topology. This has been done for some types of neuro networks like Knowledge-based networks \cite{Fu93}  \cite{Towell94}. These constitute a special class of artificial neural network that consider crude symbolic domain knowledge to generate the initial network architecture, which is later refined in the presence of training data. In the other direction there has been widespread activity aimed at translating neural language in  the form of symbolic relations \cite{Gallan88} \cite{Gallan94} \cite{Towell93}. This processes served to identify the most significant determinants of decision or classification. However this is a hard problem since often an artificial neural network with good generalization does not necessarily imply involvement of hidden units with distinct meaning. Hence any individual unit cannot essentially be associated with a single concept of feature of the problem domain. This the archetype of connectionist approaches, where all information is stored in a distributed manner among the processing units and their associated connectivity. In this work we searched for a language, based on  fuzzy logic, where the formulas are simple to inject in a multilayer feedforward network, but free from the need of given interpretation to hidden units in the problem domain.

For that we selected the language associated to a many-valued logic, the {\L}ukasiewicz logic. We inject and extract knowledge from a neural network using it. This type of logic have a very useful property motivated by the "linearity" of logic connectives. Every logic connective can be define by a neuron in an artificial network having by activation function the identity truncated to zero and one \cite{Castro98}. Allowing the direct codification of formulas in the network architecture, and simplifying the extraction of rules. This type of back-propagation neural network can be trained efficiently using the Levenderg-Marquardt algorithm, when the configuration of each neuron is conditioned to converge to predefined patters associated or having directed representation in  {\L}ukasiewicz logic.

This strategy presented good performance when applied to the reconstruction of formulas from truth tables. If the truth table is generated using a formula from the language {\L}ukasiewicz first order logic the optimum solution is defined using only units directly translated in formulas. In this type of reverse engineering problem we presuppose no noise. However the process is stable for the introduction of Gaussian noise on the input data. This motivate the application of this methodology to extract comprehensible symbolic rules from real data. However this is a hard problem since often an artificial neural network with good generalization does not necessarily imply that neural units can be translated in a symbolic formula. We describe, in this work, a simple rule to generate symbolic approximation to these unrepresentable configurations.

The presented process, for reverse engineering, can be applied to data sets characterizing a property of an entity by the truth value for a set of propositional features. And, it proved to be an excelente procedure for attribute selection. Allowing the data set simplification, by removing irrelevant attributes. The process when applied to real data generates potencial  unrepresentable models. We used the relevant inputs attributes on this models as relevante attributes to the knowledge extraction problem, deleting others. This reduces the problem dimension allowing the potencial convergence to a less complex neuronal network topology.

\textbf{Overview of the paper:}
After present the basic notions about may valued logic and how can {\L}ukasiewicz formulas be injected in a neural network. We describe the methodology for training a neural network having dynamic topology and having by activation function the identity truncated to zero and one. This methodology uses the Levenderg-Marquardt algorithm, with a special procedure called smooth crystallization to restrict the knowledge dissemination in the network structure.
The resulting configuration is pruned used a crystallization process, where only links with values near 1 or -1 survive. The complexity of the generate network is reduced by applying the "Optimal Brain Surgeon" algorithm proposed by B. Hassibi, D. G. Stork and G.J. Stork. If the simplified network doesn't satisfies the stoping criteria, the methodology is repeated in a new network, possibly with a new topology. If the data used on the network train was generated by a formula, and have sufficient cases, the process converge for a prefect connectionist presentation and every neural unit in the neural network can be reconverted to a formula. We finish this work by presenting how the describe methodology could be used to extract symbolic knowledge from real data, and how the generated model could be used as a attribute selection procedure.

\section{Preliminaries}
We begin by presenting the basic notions we need from the subjects of many valued logics,
and how formulas in its language can be injected and extracted from a back-propagations neural network.

\subsection{Many valued logic}
Classical propositional logic is one of the earliest formal systems of logic. The algebraic semantics of this logic is given by Boolean algebra. Both, the logic and the algebraic semantics have been generalized in many directions. The generalization of Boolean algebra can be based in the
relationship between conjunction and implication given by
\[
x\wedge y\leq z \Leftrightarrow x\leq y \rightarrow z \Leftrightarrow y\leq x \rightarrow z.
\]
These equivalences, called \emph{residuation equivalences}, imply the properties of logic
operators in a Boolean algebras. They can be used to present implication as a generalize inverse for the conjunction.

In application of fuzzy logic the properties of Boolean conjunction are too rigid,
hence it is extended a new binary connective $\otimes$, usually called fusion. Extending the
 commutativity to the fusion operation, the residuation equivalences define an implication
 denoted in this work by $\Rightarrow$ :
\[
x\otimes y\leq z \Leftrightarrow x\leq y \Rightarrow z \Leftrightarrow y\leq x \Rightarrow z.
\]
This two operators are supposed defined in a partially ordered set of truth values $(P,\leq)$, extending the two valued set of an Boolean algebra.  This defines a \emph{residuated poset} $(P,\otimes,\Rightarrow,\leq)$, where we interprete $P$ as a set of truth values. This structure have been used on the definition of   many types of logics. If $P$ have more than two values the associated logics are called  \emph{many-valued logics}. An infinite-valued logic is a many valued logic with $P$ infinite.

We focused our attention on many-valued logics having $[0,1]$ as set of truth values. In this
logics the fusion operator $\otimes$ is known as a \emph{t}-norm. In \cite{Gerla00} it is defined as a binary operator defined in $[0,1]$ commutative and associative, non-decreasing in both arguments and $1\otimes x= x$ and $0\otimes x= 0$.

The following are example of $t$-norms. All are continuous $t$-norms
\begin{enumerate}
  \item \emph{{\L}ukasiewicz} $t$-norm: $x\otimes y=\max(0,x+y-1)$.
  \item Product $t$-norm: $x\otimes y=xy$ usual product between real numbers.
  \item G\"{o}del $t$-norm: $x\otimes y=\min(x,y)$.
\end{enumerate}
In \cite{Frank79} all continuous $t$-norms are characterized as ordinal sums of {\L}ukasiewicz, G\"{o}del and product $t$-norms.

Many-valued logics can be conceived as a set of formal representation languages that proven to be useful for both real world and computer science applications. And when they are defined by
continuous $t$-norms they are known as \emph{fuzzy logics}.

\subsection{Processing units}

As mention in \cite{Amato02} there is a lack of a deep investigation of the relationships between logics and neural networks. In this work we present a methodology using neural networks to learn formulas from data. And where neural networks are trate as circuital counterparts of (functions represented by) formulas. They are either easy to implement and high parallel objects.

In \cite{Castro98} it is shown what, by taking as activation function $\psi$ the identity truncated to zero and one, also named saturating linear transfer function
\[
\psi(x)=\min(1,\max(x, 0))
\]

\begin{figure}
\begin{center}
 \includegraphics[width=100pt]{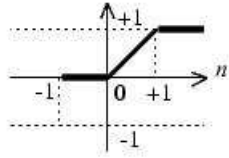}
\end{center}
\caption{Saturating linear transfer function.}\label{satlin}
\end{figure}
it is possible to represent the corresponding neural network as  combination of propositions of {\L}ukasiewicz calculus and \emph{viceversa}\cite{Amato02}.

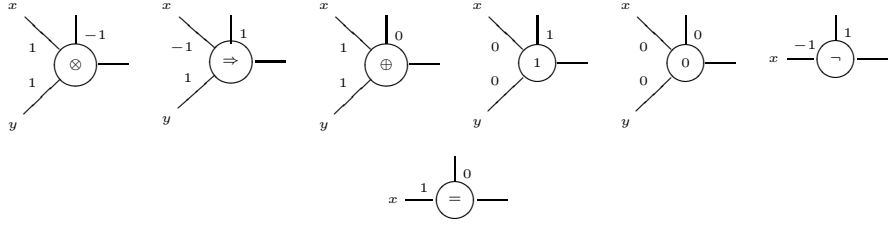
\begin{figure}[h]
\begin{center}
\tiny
$
\xymatrix @R=10pt @C=10pt { x\ar@{-}[dr]_{1} & \ar@{-}[d]^{-1} & \\
                           & *++[o][F-]{\otimes} \ar@{-}[r]& \\
           y\ar@{-}[ur]^{1} &  &  \
           }
$
$
\xymatrix @R=10pt @C=10pt { x\ar@{-}[dr]_{-1} & \ar@{-}[d]^{1} & \\
                           & *++[o][F-]{\Rightarrow} \ar@{-}[r]& \\
           y\ar@{-}[ur]^{1} &  &  \
           }
$
$
\xymatrix @R=10pt @C=10pt { x\ar@{-}[dr]_{1} & \ar@{-}[d]^{0} & \\
                           & *++[o][F-]{\oplus} \ar@{-}[r]& \\
           y\ar@{-}[ur]^{1} &  &  \
           }
$
$
\xymatrix @R=10pt @C=10pt { x\ar@{-}[dr]_{0} & \ar@{-}[d]^{1} & \\
                           & *++[o][F-]{1} \ar@{-}[r]& \\
           y\ar@{-}[ur]^{0} &  &  \
           }
$
$
\xymatrix @R=10pt @C=10pt { x\ar@{-}[dr]_{0} & \ar@{-}[d]^{0} & \\
                           & *++[o][F-]{0} \ar@{-}[r]& \\
           y\ar@{-}[ur]^{0} &  &  \
           }
$
$
\xymatrix @R=10pt @C=10pt {                    & \ar@{-}[d]^{1} & \\
           x\ar@{-}[r]^{-1} & *++[o][F-]{\neg} \ar@{-}[r]& \\
             &  &  \
           }
$
$
\xymatrix @R=10pt @C=10pt {                    & \ar@{-}[d]^{0} & \\
           x\ar@{-}[r]^{1} & *++[o][F-]{=} \ar@{-}[r]& \\
             &  &  \
           }
$
\end{center}
\caption{Neural networks codifying formulas $x\otimes y$, $x\Rightarrow y$, $x\oplus y$, $True$, $False$, $\neg x$ and $x$.}\label{neurons}
\end{figure}

Usually {\L}ukasiewicz logic sentences are built, as in first-order logic languages, from a (countable) set of propositional variables,  a conjunction $\otimes$ (the fusion operator), an implication $\Rightarrow$ and the truth constant 0.  Further connectives are defined as follows:
\begin{enumerate}
  \item $\varphi_1\wedge\varphi_2$ is $\varphi_1\otimes(\varphi_1\Rightarrow\varphi_2)$,
  \item $\varphi_1\vee\varphi_2$ is $((\varphi_1\Rightarrow\varphi_2)\Rightarrow\varphi_2)\wedge((\varphi_2\Rightarrow\varphi_1)\Rightarrow\varphi_1)$
  \item $\neg\varphi_1$ is $\varphi_1\Rightarrow 0$
  \item $\varphi_1\Leftrightarrow\varphi_2$ is $(\varphi_1\Rightarrow\varphi_2)\otimes(\varphi_2\Rightarrow\varphi_1)$
  \item $1$ is $0\Rightarrow 0$
\end{enumerate}
The usual interpretation for a well formed formula $\varphi$ is defined recursive defining by the assignment of truth values to each proposicional variable. However the application of neural network to learn {\L}ukasiewicz sentences seems more promisor using a non recursive approach to proposition evaluation. We can do this by defining the first order language  as a graphic language. In this language, words are generate using the atomic componentes presented on figure \ref{neurons}, they are networks defined linking this sort of neurons. This is made gluing atomic componentes, satisfy the neuron signature, i.e. it is an unit having several inputs and one output. This task of construct complex structures based on simplest ones can be formalized using generalized programming \cite{Fiadeiro97}.

In other words {\L}ukasiewicz logic language is defined by the set of all neural networks, where its neurons assume one of the configuration presented in figure \ref{neurons}.

\begin{figure}[h]
\begin{center}
\tiny
$
\xymatrix @R=10pt @C=10pt { x\ar@{-}[dr]_{w_1} & \ar@{-}[d]^{b} & \\
                           & *++[o][F-]{\psi} \ar@{-}[r]&z\;\;\Leftrightarrow\;\; z=\min(1,\max(0,w_1x+w_2y+b)) \\
           y\ar@{-}[ur]^{w_2} &  & \;\;=\psi_b(w_1x,w_2y) \
           }
$
\end{center}
\caption{functional interpretation for a neural network }\label{interpretation}
\end{figure}
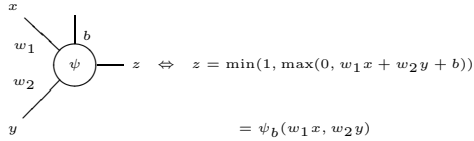

A networks of this type can be interpreted as a function, see figure \ref{interpretation}, generically denoted by $\psi_b(w_1x_1,w_2x_2)$. In this context a network is the \emph{functional interpretation} for a sentence when its interpretation is the sentence truth table.
The fact of networks and interpretation have a similar structure preserved by a graph homomorphism, called the \emph{translation morphism}, simplifies the transformation
between string base representations and the network representation, allowing to write:

\begin{prop}\label{prop1}
Every well formed formula in {\L}ukasiewicz logic language can be codified using a neural network.
\end{prop}

For instance, the semantic for sentence $\varphi=(x\otimes y\Rightarrow z)\oplus(z \Rightarrow w)$ can be described using the bellow network or codified by the bellow set of matrixes. We must note, in the example, what the partial interpretation of each unit is a simple exercise of pattern checking, relating the weights signs and the neuron bias.
\begin{center}
\tiny
\begin{tabular}{cc}
  $
\xymatrix @R=10pt @C=10pt { x\ar@{-}[dr]_{1} & \ar@{-}[d]^{-1} & \\
                           & *++[o][F-]{\otimes} \ar@{-}[rd]^{-1}& \ar@{-}[d]^{1} \\
           y\ar@{-}[ur]^{1} & *++[o][F-]{=}\ar@{-}[r]_{1}  &  *++[o][F-]{\Rightarrow} \ar@{-}[rd]_{1} &\ar@{-}[d]^{0}\\
           z\ar@{-}[dr]_{-1} \ar@{-}[ur]^{1}& \ar@{-}[d]^{1}\ar@{-}[u]_{0} & \ar@{-}[d]^{0} &*++[o][F-]{\oplus} \ar@{-}[r]&\\
                           & *++[o][F-]{\Rightarrow} \ar@{-}[r]^{1}&*++[o][F-]{=} \ar@{-}[ru]^{1}\\
           w\ar@{-}[ur]^{1} &  &  \\
           }
$ &
 \begin{tabular}{llll}

 &
 $\begin{array}{cccc}
      \;x &\; y &\; z &\; w \\
 \end{array}$ & b's & \tiny partial interpretation
      \\
$ \begin{array}{c}
      i_1  \\
      i_2 \\
      i_3 \\
    \end{array}$
    &
  $\left[
    \begin{array}{cccc}
      1 &  1 & 0 & 0 \\
      0 &  0 & 1 & 0 \\
      0 &  0 & -1 & 1\\
    \end{array}
  \right]
  $ &$ \left[
       \begin{array}{c}
         -1 \\
         0 \\
         1 \\
       \end{array}
     \right]
    $& $\begin{array}{l}
         x \otimes y \\
         z  \\
         z\Rightarrow w\\
       \end{array}$\\
    &$\begin{array}{ccc}
      \;i_1 & \;i_2 & \;i_3 \\
    \end{array}$& &
       \\
       $
  \begin{array}{c}
      j_1 \\
      j_2 \\
    \end{array}$
  &
  $\left[
    \begin{array}{ccc}
      -1 &  1 & 0 \\
       0 &  0 & 1 \\
    \end{array}
  \right]
   $&$ \left[
       \begin{array}{c}
         1 \\
         0 \\
       \end{array}
     \right]
    $& $\begin{array}{l}
         i_1\Rightarrow i_2 \\
         i_3 \\
       \end{array}$\\
   &$\begin{array}{cc}
         \;j_1 &\;j_2 \\
    \end{array}$
       \\
       &
  $\left[
       \begin{array}{cc}
         1 & 1 \\
       \end{array}
     \right] $& $\left[
       \begin{array}{c}
         0 \\
       \end{array}
     \right]$ & $j_1\oplus j_2$\\
\end{tabular} \\
& \begin{tabular}{lll}
  & &INTERPRETATION:\\
  & &$j_1\oplus j_2=(i_1\Rightarrow i_2)\oplus (i_3)=$\\
  & &$=((x \otimes y)\Rightarrow z)\oplus (z\Rightarrow w)$\\
\end{tabular}
\end{tabular}\\
\end{center}
In this sense this neural network can be seen as an interpretation for sentence $\varphi$, it codifies $f_\varphi$, the proposition truth table. And it can be presented in string base notation by writing:
\[
f_\varphi(x,y,z,w)=\psi_0(\psi_0(\psi_1(-z,w)),\psi_1(\psi_0(z),-\psi_{-1}(x,y)))
\]
However $f_\varphi$ is a continuo structure, for our propose, it must be discretized using a finite structure but having suficiente information to describe the original formula. A truth table $f_\varphi$ for a formula $\varphi$ is a map $f_\varphi:[0,1]^m\rightarrow [0,1]$, where $m$ is the number of propositional variables used in $\varphi$. For each integer $n>0$, let $S_n$ be the set $\{0,\frac{1}{n},\ldots,\frac{n-1}{n},1\}$. Each $n>0$, defines a subtable for $f_\varphi$ defined by $f_\varphi^{(n)}:(S_n)^m\rightarrow S_n$, and given by $f_\varphi^{(n)}(\bar{v})=f_\varphi(\bar{v})$, and called the $\varphi$ \emph{(n+1)-valued truth subtable}.

\subsection{Similarity between a configuration and a formula}

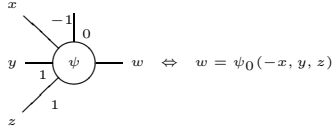
\begin{figure}
\begin{center}
\tiny
$
\xymatrix @R=10pt @C=10pt { x\ar@{-}[dr]^{-1} & \ar@{-}[d]^{0} & \\
                          y\ar@{-}[r]_{1} & *++[o][F-]{\psi} \ar@{-}[r]&w\;\;\Leftrightarrow\;\; w=\psi_0(-x,y,z) \\
           z\ar@{-}[ur]_{1} &  &  \
           }
$
\end{center}
\caption{An unrepresentable neural network}\label{unrep}
\end{figure}

We called \emph{Castro neural network} to a neural network having as activation function $\psi$ the identity truncated to zero and one and where its weights are -1, 0 or 1 and having by bias an integer. And a Carlos neural network is called \emph{representable} if it is codified as a binary neural network i.e. a  Castro neural network where each neuron don't have more than two inputs. A network is called unrepresentable if it can't be codified using a binary Castro neural network. In figure \ref{unrep}, we present an example of an unrepresentable network configuration, as we will see in the following.

Note what binary Castro neural network  can be translated directory in
the {\L}ukasiewicz first order language, and in this sense are called them {\L}ukasiewicz neural network. Bellow we presented examples of the functional interpretation for formulas with two propositional variables. They can be organized in two class:
\begin{center}
\tiny
\begin{tabular}{|c|c|}
  \hline
  & \\
  Disjunctive interpretations & Conjunctive interpretations \\
  & \\
  \hline
  $\psi_0(x_1,x_2)=f_{x_1\oplus x_2}$ & $\psi_{-1}(x_1,x_2)=f_{x_1\otimes x_2}$ \\
  $\psi_1(x_1,-x_2)=f_{x_1\oplus \neg x_2}$ & $\psi_0(x_1,-x_2)=f_{x_1\otimes \neg x_2}$ \\
  $\psi_1(-x_1,x_2)=f_{\neg x_1\oplus  x_2}$ & $\psi_0(-x_1,x_2)=f_{\neg x_1\otimes x_2}$ \\
  $\psi_2(-x_1,-x_2)=f_{\neg x_1\oplus \neg x_2}$ & $\psi_1(-x_1,-x_2)=f_{\neg x_1\otimes \neg x_2}$ \\
  \hline
\end{tabular}
\end{center}

And correspond to all possible configurations of neurons with two inputs. The other possible configurations are constant and can also be seen
as representable configurations. For instance $\psi_{b}(x_1,x_2)=0$, if $b<-1$, and
$\psi_{b}(-x_1,-x_2)=1$, if $b>1$.

In this sense every representable network can be codified by a neural network where the neural units satisfy the above patterns. Bellow we present examples of representable configurations with three inputs and how they can be codified using representable neural networks having units with two inputs.
\begin{center}
\tiny
\begin{tabular}{|c|}
  \hline
   \\
  Disjunctive configurations \\
   \\
  \hline
  $\psi_{-2}(x_1,x_2,x_3)=\psi_{-1}(x_1,\psi_{-1}(x_2,x_3))=f_{x_1\otimes x_2\otimes x_3}$ \\
  $\psi_{-1}(x_1,x_2,-x_3)=\psi_{-1}(x_1,\psi_{0}(x_2,-x_3))=f_{x_1\otimes x_2\otimes \neg x_3}$\\
  $\psi_{0}(x_1,-x_2,-x_3)=\psi_{-1}(x_1,\psi_{1}(-x_2,-x_3))=f_{x_1\otimes \neg x_2\otimes \neg x_3}$ \\
  $\psi_{1}(-x_1,-x_2,-x_3)=\psi_{0}(-x_1,\psi_{1}(-x_2,-x_3))=f_{\neg x_1\otimes \neg x_2\otimes \neg x_3}$ \\
  \hline
\end{tabular}
\end{center}

\begin{center}
\tiny
\begin{tabular}{|c|}
  \hline
   \\
 Conjunctive interpretations \\
   \\
  \hline
  $\psi_{0}(x_1,x_2,x_3)=\psi_{0}(x_1,\psi_{0}(x_2,x_3))=f_{x_1\oplus x_2\oplus x_3}$ \\
  $\psi_{1}(x_1,x_2,-x_3)=\psi_{0}(x_1,\psi_{1}(x_2,-x_3))=f_{x_1\oplus x_2\oplus \neg x_3}$ \\
  $\psi_{2}(x_1,-x_2,-x_3)=\psi_{0}(x_1,\psi_{2}(-x_2,-x_3))=f_{x_1\oplus \neg x_2\oplus \neg x_3}$ \\
  $\psi_{3}(-x_1,-x_2,-x_3)=\psi_{1}(-x_1,\psi_{2}(-x_2,-x_3))=f_{\neg x_1\oplus \neg x_2\oplus \neg x_3}$ \\
  \hline
\end{tabular}
\end{center}

Constant configurations $\psi_{b}(x_1,x_2,x_3)=0$, if $b<-2$, and
$\psi_{b}(-x_1,-x_2,-x_3)=1$, if $b>3$, are also representable. However there are example an unrepresentable network with three inputs in fig. \ref{unrep}.

Naturally, a neuron configuration when representable can by codified by
different structures using {\L}ukasiewicz neural network. Particularly we have:

\begin{prop}
If the neuron configuration $\alpha=\psi_b(x_1,x_2,\ldots,x_{n-1},x_n)$ is representable, but not constant, it can be codified in a {\L}ukasiewicz neural network  with structure: \[\beta=\psi_{b_1}(x_1,\psi_{b_2}(x_2,\ldots,\psi_{b_{n-1}}(x_{n-1},x_n)\ldots)).\]
\end{prop}

And since the $n$-nary operator $\psi_b$ is comutativa in function $\beta$ variables could interchange its position without change operator output. By this we mean what, in the string based representation, variable permutation generate equivalent formulas. From this we can concluded what:

\begin{prop}
If $\alpha=\psi_b(x_1,x_2,\ldots,x_{n-1},x_n)$ is representable, but not constant, it is the interpretation of a disjunctive formula or of a conjunctive formula.
\end{prop}

Recall that disjunctive formulas are written using only disjunctions and negations, and conjunctive formulas are written using only conjunctions and negations. This live us with the task of classify a neuron configuration according with its representation. For that, we established a relation using the configuration bias and the number of negative and positive inputs.
\begin{prop}\label{conf classification}
Given the neuron configuration
\[
\alpha=\psi_b(-x_1,-x_2,\ldots,-x_n, x_{n+1},\ldots,x_m)
\]
with $m=n+p$ inputs and where $n$ and $p$ are, respectively, the number of negative weights and the number of positive, on the neuron configuration.
\begin{enumerate}
  \item If $b=-(m-1)+n$ (i.e. $b=-p+1$) the neuron is called a \emph{conjunction} and it is a interpretation for
\[
\neg x_1\otimes\ldots\otimes\neg x_n\otimes x_{n+1}\otimes\ldots\otimes x_m
\]
  \item When $b=n$ the neuron is called a \emph{disjunction} and it is a interpretation of
\[
\neg x_1\oplus\ldots\oplus\neg x_n\oplus x_{n+1}\oplus\ldots\oplus x_m
\]
\end{enumerate}
\end{prop}

From this we proposed the following estrutural characterization for representable neurons.

\begin{prop}
Every conjunctive or disjunctive configuration $\alpha=\psi_b(x_1,x_2,\ldots,x_{n-1},x_n)$, can be codified by a  {\L}ukasiewicz
neural network \[\beta=\psi_{b_1}(x_1,\psi_{b_2}(x_2,\ldots,\psi_{b_{n-1}}(x_{n-1},x_n)\ldots)),\]
where $b=b_1+b_2+\cdots+b_{n-1}$ and $b_1\leq b_2\leq \cdots\leq b_{n-1}$.
\end{prop}

This can be translated in the following neuron rewriting rule,
\[
\tiny
\xymatrix @R=15pt @C=15pt {  \ar@{-}[rd]_{w_1} & \ar@{-}[d]^{b} &    &  \\
                      \vdots     & *++[o][F-]{\psi} \ar@{-}[r] & \ar[r]^{R} &\\
            \ar@{-}[ru]^{w_n} & &\\
           }
\xymatrix @R=15pt @C=15pt {  \ar@{-}[rd]_{w_1} & \ar@{-}[d]^{b_0} &    &  \\
                   \vdots        & *++[o][F-]{\psi} \ar@{-}[rd]^{1} & \ar@{-}[d]^{b_1}\\
            \ar@{-}[ru]^{w_{n-1}} & &*++[o][F-]{\psi}\ar@{-}[r]&\\
            \ar@{-}[rru]^{w_{n}} & &\\
           }
\]
linking networks, where values $b_0$ and $b_1$ satisfy $b=b_0+b_1$ and $b_1\leq b_0$, and such that neither involved neurons have constant output.
This rewriting rule can be used to like equivalent configurations like:
\[
\tiny
\xymatrix @R=10pt @C=10pt { x\ar@{-}[dr]^{-1} & \ar@{-}[d]^{2}                & \\
           y\ar@{-}[r]^{1} & *++[o][ F-]{\varphi} \ar@{-}[r] & \ar[r]^{R}&\\
           z\ar@{-}[ur]^{-1} &  & \\
           w\ar@{-}[uur]^{1} &  &  \\
           }
\xymatrix @R=10pt @C=10pt { x\ar@{-}[dr]^{-1} & \ar@{-}[d]^{2}                & \\
           y\ar@{-}[r]^{1} & *++[o][ F-]{\varphi} \ar@{-}[dr]^{1} & \ar@{-}[d]^{0} &\ar[r]^{R}&\\
           z\ar@{-}[ur]^{-1} &  & *++[o][ F-]{\varphi} \ar@{-}[r]&\\
           w\ar@{-}[urr]^{1} &  &  &\\
           }
\xymatrix @R=10pt @C=10pt { x\ar@{-}[dr]^{-1} & \ar@{-}[d]^{2}                  &                                &                                 &\\
                            z\ar@{-}[r]^{-1}  & *++[o][ F-]{\varphi} \ar@{-}[dr]^{1} & \ar@{-}[d]^{0}                 &                                  &\\
                            y\ar@{-}[rr]^{1}  &                                 & *++[o][ F-]{\varphi} \ar@{-}[dr]^{1}& \ar@{-}[d]^{0}                   &\\
                            w\ar@{-}[rrr]^{1} &                                 &                                & *++[o][ F-]{\varphi} \ar@{-}[r]  &\\
           }
\]
Note what a representable Castro neural network can been transformed by the application of rule R in a set of equivalente {\L}ukasiewicz neural network having less complex neurons. Then we have:
\begin{prop}
Unrepresentable neuron configurations are those transformed by
rule R in, at least, two not equivalent neural networks.
\end{prop}

For instance unrepresentable configuration $\psi_0(-x_1,x_2,x_3)$ is transform by rule R in three not equivalent configurations:
\begin{center}
\tiny
\begin{enumerate}
  \item $\psi_0(x_3,\psi_0(-x_1,x_2))=f_{x_3\oplus(\neg x_1\otimes x_2)}$,
  \item $\psi_{-1}(x_3,\psi_{1}(-x,x_2))=f_{x_3\otimes(\neg x_1\otimes x_2)}$, or
  \item $\psi_0(-x_1,\psi_0(x_2,x_3))=f_{\neg x_1\otimes(x_2\oplus x_3)}$.
\end{enumerate}
\end{center}
The representable configuration $\psi_2(-x_1,-x_2,x_3)$ is transform by rule R on only two distinct but equivalent configurations:
\begin{center}
\tiny
\begin{enumerate}
  \item $\psi_0(x_3,\psi_2(-x_1,-x_2))=f_{x_3\oplus \neg (x_1\otimes x_2)}$, or
  \item $\psi_1(-x_2,\psi_1(-x_1,x_3))=f_{\neg x_2\oplus (\neg x_1\oplus x_3)}$
\end{enumerate}
\end{center}

From this we concluded that Castro neural networks have more expressive power than {\L}ukasiewicz logic language. There are structures defined using Castro neural networks but not codified in the {\L}ukasiewicz logic language.

We also want meed reverse the knowledge injection process. We want extracted knowledge from trained neural networks. For it we need translate neuron configuration in propositional connectives or formulas. However, we as just said, not all neuron configurations can be translated in formulas, but they can be approximate by formulas. To quantify the approximation quality we defined the notion of interpretations and formulas $\lambda$-similar.

Two neuron configurations $\alpha=\psi_{b}(x_1,x_2,\ldots,x_n)$ and $\beta=\psi_{b'}(y_1,y_2,\ldots,y_n)$ are  called $\lambda$-similar
  in a $(m+1)$-valued {\L}ukasiewicz logic if $\lambda$ is the \textbf{mean absolute error} by taken the truth subtable given by $\alpha$ as an approximation to the truth subtable given by $\beta$. When this is the case we write
\[\alpha\sim_\lambda\beta.\]
If $\alpha$ is unrepresentable and $\beta$ is representable, the second configuration is called \emph{a representable approximation} to the first.

We have for instance, on the $2$-valued {\L}ukasiewicz logic (the Boolean logic case), the unrepresentable configuration $\alpha=\psi_0(-x_1,x_2,x_3)$ satisfies:
\begin{enumerate}
\tiny
  \item $\psi_0(-x_1,x_2,x_3)\sim_{0.125}\psi_0(x_3,\psi_0(-x_1,x_2))$,
  \item $\psi_0(-x_1,x_2,x_3)\sim_{0.125}\psi_{-1}(x_3,\psi_{1}(-x_1,x_2))$, and
  \item $\psi_0(-x_1,x_2,x_3)\sim_{0.125}\psi_0(-x_1,\psi_0(x_2,x_3))$.
\end{enumerate}
And in this case, the truth subtables of, formulas $\alpha_1=x_3\oplus(\neg x_1\otimes x_2)$, $\alpha_1=x_3\otimes(\neg x_1\otimes x_2)$ and $\alpha_1=\neg x_1\otimes(x_2\oplus x_3)$ are both $\lambda$-similar to $\psi_0(-x_1,x_2,x_3)$, where $\lambda=0.125$ since they differ in one position on 8 possible positions. This mean that both formulas are 12.3\% accurate. The quality of this approximations was checked by presenting values of similarity levels $\lambda$ on other finite {\L}ukasiewicz logics. For every selected logic both formulas $\alpha_1,\alpha_2$ and $\alpha_3$ have the some similarity level when compared to $\alpha$:
\begin{itemize}
\tiny
  \item $3$-valued logic,  $\lambda=0.1302$,
  \item $4$-valued logic, $\lambda=0.1300$,
  \item $5$-valued logic, $\lambda=0.1296$,
  \item $10$-valued logic, $\lambda=0.1281$,
  \item $20$-valued logic, $\lambda=0.1268$,
  \item $30$-valued logic, $\lambda=0.1263$,
  \item $50$-valued logic, $\lambda=0.1258$.
\end{itemize}

Lets see a more complex configuration $\alpha=\psi_0(-x_1,x_2,-x_3,x_4,-x_5)$. From it we can derive, through rule R, configurations:
\begin{enumerate}
\tiny
  \item $\beta_1=\psi_0(-x_5,\psi_0(x_4,\psi_0(-x_3,\psi_0(x_2,-x_1))))$
  \item $\beta_2=\psi_{-1}(x_4,\psi_{-1}(x_2,\psi_0(-x_5,\psi_0(-x_3,-x_1))))$
  \item $\beta_3=\psi_{-1}(x_4,\psi_{0}(-x_5,\psi_0(x_2,\psi_1(-x_3,-x_1))))$
  \item $\beta_4=\psi_{-1}(x_4,\psi_{0}(x_2,\psi_0(-x_5,\psi_1(-x_3,-x_1))))$
\end{enumerate}
 since this configurations are not equivalents we concluded that $\alpha$ is unrepresentable. When we compute the similarity level between $\alpha$ and each $\beta_i$ using different finite logics we have:
\begin{itemize}
 \tiny
  \item $2$-valued logic $\alpha\sim_{0.156}\beta_1$, $\alpha\sim_{0.094}\beta_2$, $\alpha\sim_{0.656}\beta_3$ and $\alpha\sim_{0.531}\beta_4$,
  \item $3$-valued logic $\alpha\sim_{0.134}\beta_1$, $\alpha\sim_{0.082}\beta_2$, $\alpha\sim_{0.728}\beta_3$ and $\alpha\sim_{0.601}\beta_4$,
  \item $4$-valued logic $\alpha\sim_{0.121}\beta_1$, $\alpha\sim_{0.076}\beta_2$, $\alpha\sim_{0.762}\beta_3$ and $\alpha\sim_{0.635}\beta_4$,
  \item $5$-valued logic $\alpha\sim_{0.112}\beta_1$, $\alpha\sim_{0.071}\beta_2$, $\alpha\sim_{0.781}\beta_3$ and $\alpha\sim_{0.655}\beta_4$,
  \item $10$-valued logic $\alpha\sim_{0.096}\beta_1$, $\alpha\sim_{0.062}\beta_2$, $\alpha\sim_{0.817}\beta_3$ and $\alpha\sim_{0.695}\beta_4$,
\end{itemize}
From this we may concluded that $\beta_2$ is a good approximation to $\alpha$ and its quality improve when we increase the number of truth values. The error increase at a low rate that the number of cases.

In this sense we will also use rule R in the case of unrepresentable configurations. From an unrepresentable configuration $\alpha$ we can generate the finite set $S(\alpha)$, with representable networks similar to $\alpha$, using rule R. Note what from $S(\alpha)$ we may select  as approximation to $\alpha$ the formula having the interpretation more similar to $\alpha$, denoted by $s(\alpha)$.  This  identification of unrepresentable configuration by representable approximations  can be used to transform network with unrepresentable neurons into representable neural networks. The stress associated to this transformation caracterizes the translation accuracy.

\subsection{A neural network crystallization}

Weights in Castro neural networks assume the values -1 or  1. However the usual learning algorithms process neural networks weights presupposing the continuity of weights domain. Naturally, every neural network with weighs in $[-1,1]$ can be seen as an approximation to a Castro neural networks. The process of identify a neural network with weighs in $[-1,1]$
with a {\L}ukasiewicz neural networks was called \emph{crystallization}. And essentially consists in rounding each neural weight $w_i$ to the nearest integer less than or equal to $w_i$, denoted by $\lfloor w_i\rfloor$.

 In this sense the crystallization process can be seen as a pruning
 on the network structure, where links between neurons with weights near 0 are removed and weights near -1 or 1 are consolidated. However this process is very crispy. We  need a smooth procedure to crystallize a network in each learning iteration to avoid the drastic reduction on learning performance.
 In each iteration we want restrict the neural network representation bias, making  the network representation bias converge to a structure similar to a Castro neural networks.
  For that, we defined by \emph{representation error} for a network $N$ with weights $w_1,\ldots,w_n$, as
 \[
 \Delta(N)=\sum^n_{i=1}(w_i-\lfloor w_i\rfloor).
 \]
 When $N$ is a Castro neural networks we have $\Delta(N)=0$. And we defined a \emph{smooth crystallization} process by iterating the function:
 \[
 \Upsilon_n(w)=sign(w).((\cos(1-abs(w)-\lfloor abs(w)\rfloor).\frac{\pi}{2})^n+\lfloor abs(w)\rfloor)
 \]
where $sign(w)$ is the sign of $w$ and $abs(w)$ its absolute value. We denote by $\Upsilon_n(N)$ the function having by input and output a neural network defined extending $\Upsilon(w_i)$ to all network weights and neurons bias. Since, for every network $N$ and $n>0$, $\Delta(N)\geq \Delta(\Upsilon_n(N))$, we have:

\begin{prop} Given a neural networks $N$ with weights in the interval $[0,1]$. For every $n>0$ the function $\Upsilon_n(N)$ have by fixed points Castro neural networks $N'$.
\end{prop}

The convergence speed dependes on parameter $n$. Increasing $n$ speedup crystallization but reduces the network plasticity to the training data.
For our applications, we selected $n=2$ based on the learning efficiency on a set of test formulas. For grater values for $n$ imposes stronger restritivos to learning. This induces a quick convergence to an admissible configuration of Castro neural network.

\section{Learning propositions}

We began the study of Castro neural network generation trying to do reverse engineering on a truth table. By this we mean what given a truth table from a $(n+1)$-valued {\L}ukasiewicz logic, \textbf{generated by a formula} in the {\L}ukasiewicz logic language, we will try to find its interpretation in the form of a {\L}ukasiewicz neural network. And from it rediscover the original formula.

For that we trained a Backpropagation neural networks using the truth table. Our methodology  trains networks having progressively more complex topologies, until a crystalized network with good performance have been found. Note that this methodology convergence dependes  on the selected training algorithm.

The bellow Algorithm \ref{RevEng} described our process for truth table reverse engineering:

\begin{algorithm}
\caption{Reverse Engineering algorithm} \label{RevEng}
\begin{algorithmic}[1]
\STATE Given a (n+1)-valued truth subtable for a {\L}ukasiewicz logic proposition
\STATE Define an inicial network complexity
\STATE Generate an inicial neural network
\STATE Apply the Backpropagation algorithm using the data set
\IF{the generated network  have bad performance}
\STATE If need increase network complexity
\STATE Try a new network. Go to 3
\ENDIF
\STATE Do neural network crystallization using the crisp process.
\IF{crystalized network have bad performance}
\STATE Try a new network. Go to 3
\ENDIF
\STATE Refine the crystalized network
\end{algorithmic}
\end{algorithm}

Given a part of a truth table we try to find a {\L}ukasiewicz neural network what codifies the data. For that we generated neural networks with a fixed number of hidden layers, on our implementation we used three. When the process detects bad learning performances, it aborts the training, and  generates a new network with random heights. After a fixed number of tries the network topology is change. This number of tries dependes of the network inputs number. After trying configure a set of networks with a given complexity and bad learning performance, the system tries to apply the selected Backpropagation algorithm to a more complex set of networks. In the following we presented a short description for the selected learning algorithm.

If the continuous optimization process converges, i.e. if the system finds a network codifying the data, the network is crystalized. If the error associated to this process increase the original network error the crystalized network is throwaway, and the system returns to the learning fase trying configure a new network.

When the process converges and the resulting network can be codified as a crisp {\L}ukasiewicz neural network the system prunes the network.  The goal of this fase is the network simplification.
For that we selected the Optimal Brain Surgeon algorithm proposed by G.J. Wolf, B. Hassibi and D.G. Stork in \cite{Hassibi93}.

\begin{figure}[h]
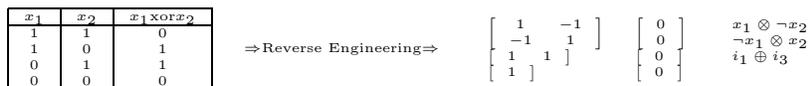

\begin{center}
\tiny
\begin{tabular}{ccc}
\begin{tabular}{|c|c|c|}
  \hline
  $x_1$ & $x_2$ & $x_1$xor$x_2$ \\
  \hline
  1 & 1 & 0 \\
  1 & 0 & 1 \\
  0 & 1 & 1 \\
  0 & 0 & 0 \\
  \hline
\end{tabular}
&
$\Rightarrow$Reverse Engineering$\Rightarrow$
&
\begin{tabular}{lll}
  $\left[
    \begin{array}{cc}
      1 &  -1 \\
      -1 &  1 \\
    \end{array}
  \right]
  $ &$ \left[
       \begin{array}{c}
         0 \\
         0 \\
       \end{array}
     \right]
    $& $\begin{array}{l}
         x_1\otimes \neg x_2 \\
         \neg x_1 \otimes  x_2  \\
       \end{array}$\\
  $\left[
    \begin{array}{cc}
      1 &  1 \\
    \end{array}
  \right]
   $&$ \left[
       \begin{array}{c}
         0 \\
       \end{array}
     \right]
    $& $\begin{array}{l}
         i_1 \oplus i_3 \\
       \end{array}$\\
  $\left[
       \begin{array}{c}
         1\\
       \end{array}
     \right] $& $\left[
       \begin{array}{c}
         0 \\
       \end{array}
     \right]$ &\\
\end{tabular}
\\
\end{tabular}
\end{center}
\caption{Input and Output structures}\label{inout}
\end{figure}

The figure \ref{inout} presents an example of the Reverse Engineering algoritmo input data set (a truth table) and output neural network structure.

\subsection{Training the neural network}
Standard Error Backpropagation algorithm (EBP) is a gradient descent algorithm, in which the network weights are moved along the negative of the gradient of the performance function. EBP algorithm has been a significant improvement in neural network research, but it has a weak convergence rate. Many efforts have been made to speed up EBP algorithm \cite{Bello92} \cite{Samad90} \cite{Solla88} \cite{Miniami90} \cite{Jacobs88}. The Levenderg-Marquardt algorithm (LM) \cite{HaganMenhaj99} \cite{Andersen95} \cite{Battiti92} \cite{Charalambous92} ensued from development of EBP algorithm dependent methods. It gives a good exchange between the speed of Newton algorithm and the stability of the steepest descent method \cite{Battiti92}.

The basic EBP algorithm adjusts the weights in the steepest descent direction. This is the direction in which the performance function is decreasing most rapidly. In the EBP algorithm, the performance index $F(w)$ to be minimized is defined as the sum of squared erros between the target output and the network's simulated outputs, namely:
\[
F(w_k)=e_k^Te_k
\]
where the vector $w_k=[w_1,w_2,\ldots,w_n]$ consists of all the current weights of the network, $e_k$ is the current error vector comprising the error for all the training examples.

When training with the EBP method, an iteration of the algorithm define the change of weights and have the form
\[
w_{k+1}=w_k-\alpha G_k
\]
where $G_k$ is the gradient of $F$ on $w_k$, and $\alpha$ is the learning rate.

Note that, the basic step of the Newton's method can be derived dom Taylor formula and is as:
\[
w_{k+1}=w_k-H_k^{-1}G_k
\]
where $H_k$ is the Hessian matrix of the performance index at the current values of the weights.

Since Newton's method implicitly uses quadratic assumptions (arising from the neglect of higher order terms in a Taylor series), The Hessian need not to be evaluated exactly. Rather an approximation can be used like
\[
H_k\approx J_k^TJ_k
\]
where $J_k$ is the Jacobian matrix that contains first derivatives of the network errors with respect to the weights $w_k$. The Jacobian matrix $J_k$ can be computed through a standard back propagation technique \cite{Mehrotra97} that is much less complex than computing the Hessian matrix. The current gradient take the form $G_k=J_k^Te_k$, where $e_k$ is a vector of current network errors. Note what $H_k= J_k^TJ_k$ in linear case. The main advantage of this technique is rapid convergence. However, the rate of convergence is sensitive to the starting location, or more precisely, the linearity around the  starting location.

It can be seen that simple gradient descent and Newton iteration are complementary in advantages they provide. Levenberg proposed an algorithm based on this observation, whose update rule is blend mentioned algorithms and is given as
\[
w_{k+1}=w_k-[J_k^TJ_k+\mu I]^{-1}J_k^Te_k
\]
where $J_k$ is the Jacobian matrix evaluated at $w_k$ and $\mu$ is the learning rate. This update rule is used as follow. If the error goes down following an update, it implies that our quadratic assumption on the function is working and we reduce $\mu$ (usually by a factor of 10) to reduce the influence of gradient descent. In this way, the performance function is always reduced at each iteration of the algorithm \cite{Megan96}. On the other hand, if the error up, we would like to follow the gradient more and so $\mu$ is increased by the same factor. The Levenberg algorithm is thus
\begin{enumerate}
  \item Do an update as directed by the rule above.
  \item Evaluated the error at the new weight vector.
  \item If error has increased as the result the update reset the weights to their previous values and increase $\mu$ by a factor $\beta$. Then try an update again
  \item If error has decreased as a result of the update, then accept the set and decrease $\mu$  by a factor $\beta$.
\end{enumerate}

The above algorithm has the disadvantage that if the value of $\mu$ is large, the approximation to Hessian matrix is not used at all. We can derive some advantage out of the second derivative even in such cases by scaling each component of the gradient according to the curvature. This should result in larger movement along the direction where the gradient is smaller so the classic "error valley" problem does not occur any more. This crucial insight was provided by Marquardt. He replaced the identity matrix in the Levenberg update rule with the diagonal of Hessian matrix approximation resulting in the  Levenberg-Marquardt update rule.
\[
w_{k+1}=w_k-[J_k^TJ_k+\mu
.diag(J_k^TJ_k)]^{-1}J_k^Te_k
\]
Since the Hessian is proportional to the curvature this rule implies a larger step in the direction with low curvatures and big step in the direction with high curvature.

The standard LM training algorithm can be illustrated in the following pseudo-codes:
\begin{algorithm}
\caption{Levenberg-Marquardt algorithm with soft crystallization} \label{LM}
\begin{algorithmic}[1]
\STATE Initialize the weights $w$ and parameters $\mu=.01$ and $\beta=.1$
\STATE Compute $e$ the sum of the squared error over all inputs $F(w)$
\STATE Compute $J$ the Jacobian of $F$ in $w$
\STATE Compute the increment of weight $\Delta w=-[J^TJ+\mu diag(J_k^TJ_k)]^{-1}J^Te$
\STATE Let $w^\ast$ be the result of applying to $w+\Delta w$ the soft crystallization process $\Upsilon_2$.
\IF{$F(w^\ast)<F(w)$}
\STATE $w=w+\Delta w$
\STATE $\mu=\mu.\beta$
\STATE Go back to step 2
\ELSE
\STATE $\mu=\mu/\beta$
\STATE Go back to step 4
\ENDIF
\end{algorithmic}
\end{algorithm}

It is to be notes while LM method is no way optimal but is just a heuristic, it works extremely well for learn {\L}ukasiewicz neural network. The only flaw is its need for matrix inversion as part of the update. Even thought the inverse is usually implemented using pseudo-inverse methods such as singular value decomposition, the cost of update become prohibitive after the model size increases to a few thousand weights.

The application of a soft cristalizador step in each iteration
accelerates the convergence to a Castro neural network.

\section{Applying reverse engineering on truth tables}

Given a {\L}ukasiewicz neural network it can be translated in the form of a string base formula if every neuron is representable. Proposition \ref{conf classification} defines a way to translate from the connectionist representation to a symbolic representation. And it is remarkable the fact that, when the truth table used in the learning is generate by a formula in a adequate   $n$-valued  {\L}ukasiewicz logic the Reverse Engineering algorithm converges to a representable {\L}ukasiewicz neural network and it is equivalent to the original formula.

When we generate a truth table in the $4$-valued  {\L}ukasiewicz logic using formula

\[
\tiny
(x_4\otimes x_5\Rightarrow x_6)\otimes(x_1\otimes x_5\Rightarrow x_2)\otimes(x_1\otimes x_2\Rightarrow x_3)\otimes(x_6\Rightarrow x_4)
\]

it have 4096 cases, the result of applying the algorithm is the 100\% accurate neural network.
\[
\tiny
\begin{tabular}{lll}
  $\left[
    \begin{array}{cccccc}
      0 &  0 & 0 & -1 & 0 & 1 \\
      0 &  0 & 0 &  1 & 1 & -1\\
      1 &  1 & -1 & 0 &  0 & 0\\
      -1 &  1 & 0 & 0 &  -1 & 0\\
    \end{array}
  \right]
  $ &$ \left[
       \begin{array}{c}
         0 \\
         -1 \\
         -1 \\
         2 \\
       \end{array}
     \right]
    $& $\begin{array}{l}
         \neg x_4\otimes x_6 \\
         x_4\otimes x_5 \otimes \neg x_6  \\
         x_1\otimes x_2 \otimes \neg x_3 \\
         \neg x_1\oplus x_2 \oplus \neg x_5
       \end{array}$\\
  $\left[
    \begin{array}{cccc}
      -1 &  -1 & -1 & 1\\
    \end{array}
  \right]
   $&$ \left[
       \begin{array}{c}
         0 \\
       \end{array}
     \right]
    $& $\begin{array}{l}
         \neg i_1 \otimes \neg i_2 \otimes \neg i_3 \otimes i_4 \\
       \end{array}$\\
  $\left[
       \begin{array}{c}
          1\\
       \end{array}
     \right] $& $\left[
       \begin{array}{c}
         0 \\
       \end{array}
     \right]$ & $j_1$\\
\end{tabular}
\]
From it we may reconstructed the formula:
\begin{center}
\tiny
$
j_1 = \neg i_1 \otimes \neg i_2 \otimes \neg i_3 \otimes i_4 =
$
$
\neg (\neg x_4\otimes x_6) \otimes \neg (x_4\otimes x_5 \otimes \neg x_6) \otimes \neg (x_1\otimes x_2 \otimes \neg x_3) \otimes (\neg x_1\oplus x_2 \oplus \neg x_5)=
$
$
= (x_4\oplus \neg x_6) \otimes (\neg x_4\oplus \neg x_5 \oplus  x_6) \otimes  (\neg x_1\oplus \neg x_2 \oplus x_3) \otimes (\neg x_1\oplus x_2 \oplus \neg x_5)=
$
$
= (x_6 \Rightarrow x_4) \otimes (x_4\otimes x_5 \Rightarrow x_6) \otimes (x_1 \otimes x_2 \Rightarrow x_3) \otimes (x_1 \otimes x_5 \Rightarrow x2)
$
\end{center}

Note however the restriction imposed, in our implementation, to three hidden layers having the least hidden layer only one neuron, impose restriction to the complexity of reconstructed formula. For instance
\[\tiny ((x_4\otimes x_5\Rightarrow x_6)\oplus(x_1\otimes x_5\Rightarrow x_2))\otimes(x_1\otimes x_2\Rightarrow x_3)\otimes(x_6\Rightarrow x_4)\]
to be codified in a three hidden layer network the last layer needs two neurons one to codify the disjunction and the other to codify the conjunctions. When the algorithm was applied to the truth table generated in the $4$-valued  {\L}ukasiewicz logic having by stoping criterium a mean square error less than $0.0007$ it produced the
representable network:
\[
\tiny
\begin{tabular}{lll}
  $\left[
    \begin{array}{cccccc}
      0 &  0 & 0 & 1 & 0 & -1 \\
      1 &  -1 & 0 &  1 & 1 & -1\\
      1 &  1 & -1 & 0 &  0 & 0\\
    \end{array}
  \right]
  $ &$ \left[
       \begin{array}{c}
         1 \\
         -2 \\
         -1 \\
       \end{array}
     \right]
    $& $\begin{array}{l}
          x_4\oplus \neg x_6 \\
         x_1\otimes \neg x_2 \otimes x_4 \otimes x_5 \otimes \neg x_6  \\
         x_1\otimes x_2 \otimes \neg x_3 \\
       \end{array}$\\
  $\left[
    \begin{array}{ccc}
      1 &  -1 & -1\\
    \end{array}
  \right]
   $&$ \left[
       \begin{array}{c}
         0 \\
       \end{array}
     \right]
    $& $\begin{array}{l}
          i_1 \otimes \neg i_2 \otimes \neg i_3 \\
       \end{array}$\\
  $\left[
       \begin{array}{c}
          1\\
       \end{array}
     \right] $& $\left[
       \begin{array}{c}
         0 \\
       \end{array}
     \right]$ & $j_1$\\
\end{tabular}
\]
By this we may conclude what original formula can be approximate, or is $\lambda$-similar with $\lambda=0.002$ to:
\begin{center}
\tiny
$j_1=i_1 \otimes \neg i_2 \otimes \neg i_3=$
$( x_4\oplus \neg x_6 ) \otimes \neg ( x_1\otimes \neg x_2 \otimes x_4 \otimes x_5 \otimes \neg x_6) \otimes \neg (x_1\otimes x_2 \otimes \neg x_3)=$
$=( x_4\oplus \neg x_6 ) \otimes (\neg x_1\oplus  x_2 \oplus \neg x_4 \oplus \neg x_5 \oplus x_6) \otimes (\neg x_1\oplus \neg x_2 \oplus x_3)=$
$=( x_6 \Rightarrow x_4) \otimes ((x_1 \otimes x_4\otimes x_5) \Rightarrow (x_2\oplus x_6) ) \otimes (x_1\otimes x_2 \Rightarrow x_3)$
\end{center}
Note that $j_1$ is 0.002-similar to the original formula in the $4$-valued  {\L}ukasiewicz logic but it is equivalente to the original in the $2$-valued  {\L}ukasiewicz logic, i.e. in Boolean logic.

The fixed number of layer also impose restrictions to reconstruction of formula. A table generated by:
\[\tiny (((i_1\otimes i_2) \oplus (i_2 \otimes i_3)) \otimes ((i_3\otimes i_4)\oplus (i_4\otimes i_5))) \oplus (i_5\otimes i_6)\]
requires at least 4 hidden layers, to be reconstructed, this is the number os levels required by the associated parsing tree.

Bellow we can see all the fixed points found by the process, when applied on the 5-valued truth table for \[x\wedge y:=\min(x,y).\]
These reversed formulas are equivalent in the $5$-valued  {\L}ukasiewicz logic, and where find for different executions.
\[
\tiny
\begin{tabular}{cccc}
  \xymatrix @R=10pt @C=10pt {                  & \ar@{-}[d]^{0}                & \ar@{-}[d]^{-1} &\\
           y\ar@{-}[r]^{-1} & *++[o][F-]{\varphi} \ar@{-}[r]^{-1} & *++[o][F-]{\varphi}\ar@{-}[r]&\\
           x\ar@{-}[ur]^{1}\ar@{-}[r]^{-1} & *++[o][F-]{\varphi} \ar@{-}[ru]^{-1}& &\\
                            & \ar@{-}[u]^{1} &  \
           } & \xymatrix @R=10pt @C=10pt {                  & \ar@{-}[d]^{0}                & \ar@{-}[d]^{0} &\\
           y\ar@{-}[r]^{-1} & *++[o][F-]{\varphi} \ar@{-}[r]^{-1} & *++[o][F-]{\varphi}\ar@{-}[r]&\\
           x\ar@{-}[ur]^{1}\ar@{-}[r]^{1} & *++[o][F-]{\varphi} \ar@{-}[ru]^{1}& &\\
                            & \ar@{-}[u]^{0} &  \
           } & \xymatrix @R=10pt @C=10pt {                  & \ar@{-}[d]^{0}                & \ar@{-}[d]^{0} &\\
           y\ar@{-}[r]^{1} & *++[o][F-]{\varphi} \ar@{-}[r]^{-1} & *++[o][F-]{\varphi}\ar@{-}[r]&\\
           x\ar@{-}[ur]^{-1}\ar@{-}[r]^{1} & *++[o][F-]{\varphi} \ar@{-}[ru]^{1}& &\\
                            & \ar@{-}[u]^{0} &  \
           } & \xymatrix @R=10pt @C=10pt {                  & \ar@{-}[d]^{1}                & \ar@{-}[d]^{0} &\\
           y\ar@{-}[r]^{1} & *++[o][F-]{\varphi} \ar@{-}[r]^{1} & *++[o][F-]{\varphi}\ar@{-}[r]&\\
           x\ar@{-}[ur]^{-1}\ar@{-}[r]^{-1} & *++[o][F-]{\varphi} \ar@{-}[ru]^{-1}& &\\
                            & \ar@{-}[u]^{1} &  \
           } \\
  $\neg(\neg\neg(x\Rightarrow y)\Rightarrow \neg x)$ & $\neg(x\Rightarrow \neg(x\Rightarrow y))$ & $\neg(\neg(y\Rightarrow x)\Rightarrow x)$ & $\neg((x\Rightarrow y)\Rightarrow \neg x)$ \\
\end{tabular}
\]
\[
\tiny
\begin{tabular}{cccc}
 \xymatrix @R=10pt @C=10pt {                  & \ar@{-}[d]^{0}                & \ar@{-}[d]^{0} &\\
           x\ar@{-}[r]^{-1} & *++[o][F-]{\varphi} \ar@{-}[r]^{-1} & *++[o][F-]{\varphi}\ar@{-}[r]&\\
           y\ar@{-}[ur]^{1}\ar@{-}[r]^{1} & *++[o][F-]{\varphi} \ar@{-}[ru]^{1}& &\\
                            & \ar@{-}[u]^{0} &  \
           }  & \xymatrix @R=10pt @C=10pt {                  & \ar@{-}[d]^{0}                & \ar@{-}[d]^{-1} &\\
           x\ar@{-}[r]^{1} & *++[o][F-]{\varphi} \ar@{-}[r]^{1} & *++[o][F-]{\varphi}\ar@{-}[r]&\\
           y\ar@{-}[ur]^{1}\ar@{-}[r]^{-1} & *++[o][F-]{\varphi} \ar@{-}[ru]^{1}& &\\
                            & \ar@{-}[u]^{1} &  \
           } & \xymatrix @R=10pt @C=10pt {                  & \ar@{-}[d]^{1}                & \ar@{-}[d]^{1} &\\
           x\ar@{-}[r]^{-1} & *++[o][F-]{\varphi} \ar@{-}[r]^{-1} & *++[o][F-]{\varphi}\ar@{-}[r]&\\
           y\ar@{-}[ur]^{-1}\ar@{-}[r]^{1} & *++[o][F-]{\varphi} \ar@{-}[ru]^{-1}& &\\
                            & \ar@{-}[u]^{0} &  \
           } & \xymatrix @R=10pt @C=10pt {                  & \ar@{-}[d]^{1}                & \ar@{-}[d]^{0} &\\
           x\ar@{-}[r]^{1} & *++[o][F-]{\varphi} \ar@{-}[r]^{1} & *++[o][F-]{\varphi}\ar@{-}[r]&\\
           y\ar@{-}[ur]^{-1}\ar@{-}[r]^{-1} & *++[o][F-]{\varphi} \ar@{-}[ru]^{-1}& &\\
                            & \ar@{-}[u]^{1} &  \
           } \\
  $\neg(y\Rightarrow \neg(y\Rightarrow x))$ & $(y\Rightarrow x)\otimes y$ & $\neg(\neg(y\Rightarrow x)\Rightarrow \neg y)$ & $\neg((y\Rightarrow x)\Rightarrow\neg y)$ \\
\end{tabular}
\]

The bellow table presents mean times need to find a configuration with a mean square error less than 0.002. Then mean time is computed using a 6 tries for some formulas on the 5-valued truth  {\L}ukasiewicz logic. We implementation the algorithm using the MatLab neural network package and run it in a AMD Athalon 64 X2 Dual-Core Processor TK-53 at 1.70 GH on a Windows Vista system with 1G of  memory.
\begin{center}
\tiny
\begin{tabular}{|c|r||r|r|}
  \hline
  &formula & mean & variance \\
  \hline
  1& $i_1\otimes i_3\Rightarrow i_6$ & 5.68 & 39.33 \\
  2& $i_4\Rightarrow i_6\otimes i_6\Rightarrow i_2$ & 26.64 & 124.02 \\
  3& $((i_1\Rightarrow i_4)\oplus(i_6\Rightarrow i_2))\otimes(i_6\Rightarrow i_1)$ & 39.48 & 202.94 \\
  4& $(i_4\otimes i_5\Rightarrow i_6)\otimes(i_1\otimes i_5\Rightarrow i_2)$ & 51.67 & 483.85 \\
  5& $((i_4\otimes i_5\Rightarrow i_6)\oplus(i_1\otimes i_5\Rightarrow i_2))\otimes(i_1\otimes i_3\Rightarrow i_2)$ & 224.74 & 36475.47 \\
  6& $((i_4\otimes i_5\Rightarrow i_6)\oplus(i_1\otimes i_5\Rightarrow i_2))\otimes(i_1\otimes i_3\Rightarrow i_2)\otimes(i_6\Rightarrow i_4)$ & 368.32 & 55468.66 \\
  \hline
\end{tabular}
\end{center}

\section{Applying the process on real data}

The extraction of a rule from a data set is very different from the task of reverse engineering the rule used on the generation of a data set. In sense what, in the reverse engineering task we know the existence of a prefect description for the information, we know the adequate logic language to describe it and we have lack of noise. The extraction of a rule from a data set is made establishing a stopping criterium base on a language fixed by the extraction process. The expressive power of this language caracterize the learning algorithm plasticity. However very expressive languages produce good fitness to the trained data, but with bad generalization, and its sentences are usually difficult to understand.

With the application of our process to real data we try to catch information in the data similar to the information described using sentences in {\L}ukasiewicz logic language. This naturally means what, in this case, we will try to search for simple and understandable models for the data. And for this make sense strategy followed of train of progressively more complex models and subjected to a strong criteria of pruning. When the mean squared error stopping criteria is satisfied it has big probability of be the simplest one. However some of its neuron configurations may be unrepresentable and must be approximated by a formula without damage drastically the model performance.

Note however the fact what the use of the presented process can be prohibitive to train complex models having a grate number of attributes, i.e. learn formulas with many connectives and propositional variables. In this sense our process use must be preceded by a fase of attribute selection.

\subsubsection{Mushrooms}
\emph{Mushroom} is a data set available in \emph{UCI Machine Learning Repository}. Its records drawn from The Audubon Society Filed Guide to North American Mushrooms (1981) G. H. Lincoff (Pres.), New York, was donate by Jeff Schlimmer. This data set includes descriptions of hypothetical samples corresponding to 23 species of gilled mushrooms in the Agaricus and Lepiota Family. Each species is identified as definitely edible, definitely poisonous, or of unknown edibility and not recommended. This latter class was combined with the poisonous one. The Guide clearly states that there is no simple rule for determining the edibility of a mushroom. However we will try to find a one using the data set as a truth table.

The data set have 8124 instances defined using 22 nominally valued attributes presented in the table bellow. It has missing attribute values, 2480, all for attribute \#11. 4208 instances (51.8\%) are classified as editable and 3916 (48.2\%) has classified poisonous.
\begin{center}
\tiny
\begin{tabular}{|l|l|l|}
  \hline
  N. & Attribute &Values \\
  \hline
  0& classes & edible=e, poisonous=p \\
  1& cap.shape & bell=b,conical=c,convex=x,flat=f,knobbed=k,sunken=s \\
  2& cap.surface & fibrous=f,grooves=g,scaly=y,smooth=s \\
  3& cap.color & brown=n,buff=b,cinnamon=c,gray=g,green=r,pink=p,purple=u,red=e,white=w,\\
   && yellow=y  \\
  4& bruises? & bruises=t,no=f \\
  5& odor & almond=a,anise=l,creosote=c,fishy=y,foul=f,musty=m,none=n,pungent=p,\\
   &&spicy=s \\
  6& gill.attachment & attached=a,descending=d,free=f,notched=n \\
  7& gill.spacing & close=c,crowded=w,distant=d \\
  8& gill.size & broad=b,narrow=n \\
  9& gill.color & black=k,brown=n,buff=b,chocolate=h,gray=g,green=r,orange=o,pink=p,\\
  &&purple=u,red=e,white=w,yellow=y \\
  10& stalk.shape & enlarging=e,tapering=t \\
  11& stalk.root & bulbous=b,club=c,cup=u,equal=e,rhizomorphs=z,rooted=r,missing=? \\
  12& stalk.surface.above.ring & ibrous=f,scaly=y,silky=k,smooth=s \\
  13& stalk.surface.below.ring & ibrous=f,scaly=y,silky=k,smooth=s \\
  14& stalk.color.above.ring & brown=n,buff=b,cinnamon=c,gray=g,orange=o,pink=p,red=e,white=w,yellow=y \\
  15& stalk.color.below.ring & brown=n,buff=b,cinnamon=c,gray=g,orange=o,pink=p,red=e,white=w,yellow=y \\
  16& veil.type & partial=p,universal=u \\
  17& veil.color & brown=n,orange=o,white=w,yellow=y \\
  18& ring.number & none=n,one=o,two=t \\
  19& ring.type & cobwebby=c,evanescent=e,flaring=f,large=l,none=n,pendant=p,sheathing=s,\\
  &&zone=z \\
  20& spore.print.color & black=k,brown=n,buff=b,chocolate=h,green=r,orange=o,purple=u,white=w,\\
  &&yellow=y \\
  21& population & abundant=a,clustered=c,numerous=n,scattered=s,several=v,solitary=y \\
  22& habitat & grasses=g,leaves=l,meadows=m,paths=p,urban=u,waste=w,woods=d \\
  \hline
\end{tabular}
\end{center}

We used a unsupervised filter converting all nominal attributes into binary
numeric attributes. An attribute with $k$ values is transformed into $k$
binary attributes if the class is nominal. This produces a data set with
111 binary attributes.

After the binarization we used the presented algorithm to selected relevante attributes for mushrooms classification. After 4231.8 seconds the system
produced a model, having an architecture (2,1,1), a quite complex rule
with 100\% accuracy depending  on 23 binary attributes defined by values of
\begin{center}
\tiny
\{odor,gill.size,stalk.surface.above.ring, ring.type, spore.print.color\}
\end{center}
With the values assumed by this attributes we produce a new data set. After
some tries the simples model generated was the following:
\[
\tiny
\xymatrix @R=10pt @C=10pt
     {  A1- bruises? = t \ar@{-}[dddrrr]^{1}&&&  & \\
        A2- odor\in\{a,l,n\}\ar@{-}[ddrrr]^{1} &&& \ar@{-}[dd]^{1} & \\
        A3- odor=c \ar@{-}[drrr]^{-1}&&&  & \\
        A4- ring.type = e \ar@{-}[rrr]^{-1}   &&& *++[o][F-]{\varphi} \ar@{-}[r]& \\
        A5- spore.print.color = r \ar@{-}[urrr]^{-1}&&&  & \\
        A6- population = c \ar@{-}[uurrr]^{-1}&&&  & \\
        A7- habitat = w\ar@{-}[uuurrr]^{1} &&&  & \\
        A8- habitat \in \{g,m,u,d,p,l\} \ar@{-}[uuuurrr]_{-1}&&&  &  \
     }
\]

This model have an accuracy of 100\%. From it, and since attribute values in A2 and A3, and in A7 and A8 are
auto exclusive, we used propositions A1, A2, A3, A4, A5, A6, A7 to define a new data set. This new data set was
enriched with new negative cases by introduction for each original case a new one where the
truth value of each attribute was multiplied by 0.5. For instance the "eatable" mushroom case:
\begin{center}
\tiny
 (A1=0, A2=1, A3=0, A4=0, A5=0, A6=0, A7=0,A8=1,A9=0)
\end{center}
was used on the definition of a new "poison" case
\begin{center}
\tiny
 (A1=0, A2=0.5, A3=0, A4=0, A5=0, A6=0, A7=0,A8=0.5,A9=0)
\end{center}
This resulted in a convergence speed increase and reduced the occurrence of no representable
configurations.

When we applied our "reverse engineering" algoritmo to the enriched data set, having by stoping criteria the mean square error less than \emph{mse}. For $mse=0.003$ the system produced the model:
\[
\tiny
\begin{tabular}{lll}
  $\left[
    \begin{array}{ccccccc}
      0 &  1 & 0 & 0 & -1 & 0 & 1 \\
      0 &  1 & 0 & 1 &  0 & 0 & -1 \\
    \end{array}
  \right]
  $ &$ \left[
       \begin{array}{c}
         -1 \\
         -1 \\
       \end{array}
     \right]
    $& $\begin{array}{l}
         A2\otimes \neg A5 \otimes A7 \\
         A2\otimes A4\otimes \neg A7  \\
       \end{array}$\\
  $\left[
    \begin{array}{cc}
      1 & 1 \\
    \end{array}
  \right]
   $&$ \left[
       \begin{array}{c}
         0 \\
       \end{array}
     \right]
    $& $i_1\oplus i_2$\\
  $\left[
       \begin{array}{c}
         1 \\
       \end{array}
     \right] $& $\left[
       \begin{array}{c}
         0 \\
       \end{array}
     \right]$ \\
\end{tabular}
\]

This model codifies the proposition
\begin{center}
\tiny $(A2\otimes \neg A5 \otimes A7)\oplus(A2\otimes A4\otimes \neg A8)$
\end{center}
and misses the classification of 48 cases. It have 98.9\% accuracy.

More precise model can be produced, by restring the stoping criteria. However this, in general, produce more complex propositions and more dificulte to understand. For instance with $mse=0.002$ the systems generated the bellow model. It misses 32 cases, having an accuracy of 99.2\%, and easy to convert in a proposition.
\[
\tiny
\begin{tabular}{lll}
  $\left[
    \begin{array}{ccccccc}
      0 &  0 & 0 & -1 & 0 & 0 & 1 \\
      1 &  1 & 0 & -1 &  0 & 0 & 0 \\
      0 &  0 & 0 & 0 &  0 & 0 & 1 \\
      0 &  1 & 0 & 0 &  -1 & -1 & 1 \\
    \end{array}
  \right]
  $ &$ \left[
       \begin{array}{c}
         1 \\
         -1 \\
         0 \\
         -1\\
       \end{array}
     \right]
    $& $\begin{array}{l}
         \neg A4 \oplus A7 \\
         A1\otimes A2\otimes \neg A4  \\
         A7\\
         A2\otimes \neg A5\otimes \neg A6\otimes A7\\
       \end{array}$\\
  $\left[
    \begin{array}{cccc}
      -1 &  0 & 1 &  0 \\
       1 & -1 & 0 & -1 \\
    \end{array}
  \right]
   $&$ \left[
       \begin{array}{c}
         1 \\
         0 \\
       \end{array}
     \right]
    $& $\begin{array}{l}
         \neg i_1\oplus i_3 \\
         i_1\otimes \neg i_2 \otimes \neg i_4 \\
       \end{array}$\\
  $\left[
       \begin{array}{cc}
         1 & -1 \\
       \end{array}
     \right] $& $\left[
       \begin{array}{c}
         0 \\
       \end{array}
     \right]$ & $j_1\otimes\neg j_2$\\
\end{tabular}
\]
This neural network codifies
\begin{center}
\tiny
$j_1\otimes\neg j_2 =
(\neg i_1\oplus i_3)\otimes \neg(i_1\otimes \neg i_2 \otimes \neg i_4)=$
$=(\neg (\neg A4 \oplus A7)\oplus A7)\otimes \neg((\neg A4 \oplus A7)\otimes \neg (A1\otimes A2\otimes \neg A4) \otimes \neg (A2\otimes \neg A5\otimes \neg A6\otimes A7))=
$
$
=((A4 \otimes \neg A7)\oplus A7)\otimes (( A4 \otimes \neg A7)\oplus (A1\otimes A2\otimes \neg A4) \oplus (A2\otimes \neg A5\otimes \neg A6\otimes A7))
$
\end{center}

Some times the algorithm converged to unrepresentable configurations like the one presented bellow, having however 100\% accuracy. The frequency of this type of configurations increases with the increase of required accuracy.
\[
\tiny
\begin{tabular}{lll}
  $\left[
    \begin{array}{ccccccc}
      -1 &  1 & -1 & 1 & 0 & -1 & 0 \\
      0 &  0 & 0 & 1 &  1 & 0 & -1 \\
      1 &  1 & 0 & 0 &  0 & 0 & -1 \\
    \end{array}
  \right]
  $ &$ \left[
       \begin{array}{c}
         0 \\
         1 \\
         0 \\
       \end{array}
     \right]
    $& $\begin{array}{l}
         i_1\text{ unrepresentable} \\
         A4\otimes A5\otimes \neg A6  \\
         i_3\text{ unrepresentable} \\
       \end{array}$\\
  $\left[
    \begin{array}{ccc}
      1 &  -1 & 1 \\
    \end{array}
  \right]
   $&$ \left[
       \begin{array}{c}
         0 \\
       \end{array}
     \right]
    $& $\begin{array}{l}
         j_1 \text{unrepresentable} \\
       \end{array}$\\
  $\left[
       \begin{array}{c}
         1 \\
       \end{array}
     \right] $& $\left[
       \begin{array}{c}
         0 \\
       \end{array}
     \right]$ & \\
\end{tabular}
\]
Since, for the similarity evaluation on data set, we have:
\begin{enumerate}
\tiny
  \item $i_1\sim_{0.0729}((\neg A1\otimes A4) \oplus A2)\otimes \neg A3 \otimes \neg A6$
  \item $i_3\sim_{0.0}(A1\oplus\neg A7)\otimes A2$
  \item $j_1\sim_{0.0049}(i_1\otimes\neg i_2)\oplus i_3$
\end{enumerate}
The formula
\begin{center}
\tiny
$\alpha=(((((\neg A1\otimes A4) \oplus A2)\otimes \neg A3 \otimes \neg A6)\otimes\neg (A4\otimes A5\otimes \neg A6))\oplus ((A1\oplus\neg A7)\otimes A2)$
\end{center}
is $\lambda$-similar, with $\lambda=0.0049$ to the original neural network. Formula $\alpha$ misses the classification for 40 cases. Note what the symbolic model is stable to the bad performance of $i_1$ representation.

Other example of unrepresentable is given bellow. This network structure can be simplified during the symbolic translation.
\[
\tiny
\begin{tabular}{lll}
  $\left[
    \begin{array}{ccccccc}
      1 &  1 & -1 & 1 & 0 & 0 & 1 \\
      0 &  0 & 1 & -1 &  0 & 0 & 0 \\
      0 &  -1 & 0 & 1 &  1 & 0 & -1 \\
    \end{array}
  \right]
  $ &$ \left[
       \begin{array}{c}
         -1 \\
         0 \\
         2 \\
       \end{array}
     \right]
    $& $\begin{array}{l}
         i_1\text{ unrepresentable} \\
         A3\otimes \neg A4  \\
         \neg A2\otimes A4\otimes A5 \otimes \neg A7 \\
       \end{array}$\\
  $\left[
    \begin{array}{ccc}
      -1 &  0 & 1 \\
      0 &   1 & 1 \\
    \end{array}
  \right]
   $&$ \left[
       \begin{array}{c}
         0 \\
         1 \\
       \end{array}
     \right]
    $& $\begin{array}{l}
         \neg i_1 \otimes i_3 \\
         1 \\
       \end{array}$\\
  $\left[
       \begin{array}{cc}
         -1 & 1\\
       \end{array}
     \right] $& $\left[
       \begin{array}{c}
         1 \\
       \end{array}
     \right]$ & $\neg j_1 \otimes j_2$\\
\end{tabular}
\]
Since
\[
\tiny
i_1\sim_{0.0668}(A1\otimes A2 \otimes A7)\oplus \neg A3 \oplus A4
\]
the neural network is similar to,
\begin{center}
\tiny
$\alpha=\neg j_1 \otimes j_2=\neg (\neg i_1 \otimes i_3) \otimes 1 =$
$((A1\otimes A2 \otimes A7)\oplus \neg A3 \oplus A4)\oplus\neg (\neg A2\otimes A4\otimes A5 \otimes \neg A7)$
\end{center}
and the degree of similarity is $\lambda=0$, i.e. the neural network interpretation is equivalent to formula $\alpha$ in the Mushrooms data set, in the sense what both produce equal classifications.

\section{Conclusions}
This methodology to codify and extract symbolic knowledge from a neuro network is very simple and efficient for the extraction of simple rules from medium sized data sets. From our experience  the described algorithm is a very good tool for attribute selection, particulary when we have low noise and classification problems depending from few nominal attributes to be selected from a huge set of possible attributes.

In the theoretic point of view it is particularly interesting the fact what restricting the values assumed by neurons weights restrict the information propagation in the network. Allowing the emergence of patterns in the neuronal network structure. For the case of linear neuronal networks these structures are characterized by the occurrence of patterns in neuron configuration with a direct symbolic presentation in a {\L}ukasiewicz logic.

\bibliographystyle{amsplain}
\bibliography{aveiro}
\end{document}